# Multi-Dimensional AGV Path Planning in 3D Warehouses Using Ant Colony Optimization and Advanced Neural Networks

Bo Zhang[1], Xiubo Liang*, Wei Song[2], Yulu Chen[2]

[1] School of Software Technology, Zhejiang University, Ningbo 315100, China
[2] Loctek Ergonomic Technology Co., Ningbo 315100, China
Corresponding author: Xiubo Liang (xiubo@zju.edu.cn)

**Abstract.** Within modern warehouse scenarios, the rapid expansion of e-commerce and increasingly complex, multi-level storage environments have exposed the limitations of traditional AGV (Automated Guided Vehicle) path planning methods—often reliant on static 2D models and expert-tuned heuristics that struggle to handle dynamic traffic and congestion. Addressing these limitations, this paper introduces a novel AGV path planning approach for 3D warehouse environments that leverages a hybrid framework combining ACO (Ant Colony Optimization) with deep learning models, called NAHACO (Neural Adaptive Heuristic Ant Colony Optimization). NAHACO integrates three key innovations: first, an innovative heuristic algorithm for 3D warehouse cargo modeling using multi-dimensional tensors, which addresses the challenge of achieving superior heuristic accuracy; second, integration of a congestion-aware loss function within the ACO framework to adjust path costs based on traffic and capacity constraints, called CARL (Congestion-Aware Reinforce Loss), enabling dynamic heuristic calibration for optimizing ACO-based path planning; and third, an adaptive attention mechanism that captures multi-scale spatial, thereby addressing dynamic heuristic calibration for further optimization of ACO-based path planning and AGV navigation. NAHACO significantly boosts path planning efficiency, yielding faster computation times and superior performance over both vanilla and state-of-the-art methods, while automatically adapting to warehouse constraints for real-time optimization. NAHACO outperforms state-of-the-art methods. It lowers the total cost by up to 24.7% on TSP benchmarks. In warehouse tests, NAHACO cuts cost by up to 41.5% and congestion by up to 56.1% compared to previous methods.

**Keywords:** 3D Warehouse Path PlanningAnt Colony OptimizationDeep LearningNeural NetworksCargo ModelingCongestion MitigationNeural-Enhanced Meta-Heuristic

## 1 Introduction

AGV (Automated Guided Vehicle) is integral to contemporary warehouse automation, particularly as the rapid expansion of e-commerce transforms storage facilities into sophisticated, multi-level 3D environments where operational efficiency, energy conservation, and safety are paramount. In such complex settings, AGV not only facilitates seamless goods transportation but also serve as crucial nodes in a larger, interconnected digital ecosystem, driving real-time decision making and system-wide optimization. Among the advanced computational strategies adopted to enhance AGV navigation, ACO (Ant Colony Optimization) stands out for its bio-inspired approach that effectively navigates vast, dynamic search spaces by mimicking the collective foraging behavior of ants. This adaptable algorithm has been widely embraced for AGV path planning, as it offers a promising means to address the intricacies of multi-level 3D warehouse configurations and overcome the limitations of static, heuristic-based methods.

Despite the historical success of traditional path planning methodologies for AGV, their reliance on static 2D representations and expert-tuned heuristics has rendered them increasingly inadequate for the demands of modern, multi-level 3D warehouse environments. These vanilla methods are challenged by the need to navigate complex spatial configurations that involve dynamic obstacles, variable cargo attributes—such as differing dimensions, weights, and specialized handling requirements—and the fluctuating patterns of real-time congestion. As a result, the inability of traditional models to capture such multifaceted interactions leads to suboptimal routing decisions, diminished operational efficiency, and elevated safety risks in high-traffic conditions. Although classical implementations of ACO have demonstrated a degree of robustness and adaptability in controlled scenarios, they often struggle to address the nuanced challenges presented by intricate 3D environments. In response, recent research efforts have focused on integrating advanced deep learning models with ACO, with several studies proposing hybrid approaches that incorporate deep reinforcement learning to enhance path planning capabilities in multi-task AGV scenarios [36, 38, 39]. Despite these promising advances, such integrations tend to incur significant computational costs and still fall short of fully mitigating AGV congestion issues or comprehensively accounting for the diverse characteristics of cargo in 3D spaces. Consequently, there remains a pressing need for further innovation to develop more resilient and scalable solutions that can effectively meet the complex operational requirements of contemporary warehouse systems.

To address these critical challenges, this paper introduces an advanced hybrid framework that synergistically combines an enhanced ACO algorithm with state-of-the-art deep learning techniques to revolutionize AGV path planning in sophisticated 3D warehouse environments. Central to our methodology is the development of an innovative heuristic algorithm that leverages multi-dimensional tensor representations, thereby enabling the detailed modeling of intricate spatial and physical cargo characteristics—ranging from precise dimensions and weight distributions to specific handling requirements. This refined modeling strategy not only enhances the granularity of the path planning process but also facilitates a comprehensive reflection of the dynamic environmental constraints prevalent in modern warehouses. Moreover, our framework incorporates a dynamically adaptive loss function within the ACO paradigm, called CARL (Congestion-Aware Reinforce Loss), which empowers the algorithm to adjust path costs in real time by effectively accounting for fluctuating traffic densities and capacity limitations, thus mitigating congestion and ensuring more optimal routing decisions. It employs a neural network framework that integrates GNN (Graph Neural Network) for spatial feature extraction, an attention-based model for static-dynamic feature fusion module, and an MLP (Multi-Layer Perceptron) for heuristic decoding. This architecture enables the dynamic generation of heuristic information, replacing static expert-defined heuristics with adaptive, data-driven insights. By leveraging real-time warehouse data, the system refines heuristic values to enhance ACO-based path planning, optimizing AGV navigation in complex 3D environments. Our approach introduces a customized fusion mechanism that improves spatiotemporal feature extraction and a novel loss function designed to optimize the congestion level of AGV paths. This allows for more precise predictions of essential operating parameters, significantly improving decision accuracy in dynamic warehouse environments. Loctek's tests confirm our NAHACO method. NAHACO reduces transportation cost by up to 47% compared to state-of-the-art methods. It lowers congestion by up to 41% relative to current solutions. These results prove that NAHACO is a scalable and efficient solution for modern warehouse systems.

## 2  Related Work

### 2.1  Neural Combinatorial Optimization

NCO (Neural Combinatorial Optimization) combines neural networks with reinforcement learning to solve the CO (Combinatorial Optimization) problem. In the first category, end-to-end methods are used to acquire knowledge about the construction of autoregressive solutions or heatmap generation, which are then used in the subsequent sampling-based decoding process. In this area, recent progress has been reflected in multiple directions. Specifically, some neural architectures have achieved higher levels of alignment [1, 2, 3, 4, 5]. Training paradigms have been upgraded to more complex levels [6, 7, 8]. Solution pipelines have evolved into more advanced forms [9, 10, 11]. The scope of applications has also expanded to a wider range [12, 13, 14]. End-to-end methods are indeed very efficient. Bello et al. [15] proposed NCO, which was first applied to the TSP (Travelling Salesman Problem), demonstrating the potential of learned solutions over traditional heuristics. Subsequent studies, including Garmendia et al. [16] and Liu et al. [17], evaluated its effectiveness and limitations, particularly for the TSP, while Verdù et al. [18] improved the results using simulation-guided beam search. NCO has also been tailored for real-world applications such as online vehicle routing [19], and Wu et al. [20] conducted a comprehensive survey of its application to the vehicle routing problem. Efforts to improve scalability by Luo et al. [21, 22] and others have addressed large-scale CO instances, though Garmendia et al. [23] caution against context-specific weaknesses.

## 2.2 Cargo Loading Optimization in Warehouse Environments and Ant Colony Optimization for AGV Picking Path Planning

Recent cutting-edge cargo loading optimization heuristics combine greedy and tabu search algorithms with 3D loading constraints for periodic pickup and delivery, such as the greedy-tabu dual heuristic algorithm by Xu et al. [24], while Ananno et al. [25] proposed a multi-heuristic packing algorithm that combines constructive heuristics and genetic optimization to improve stability, space utilization, and practical applicability, thereby improving driving efficiency, load stability, and reducing material waste.

In parallel, ACO, inspired by the foraging behavior of real ant colonies, has emerged as an efficient metaheuristic for approximate optimization. ACO is a metaheuristic for approximate optimization, first proposed by Dorigo et al. [26]. ACO employs indirect communication through pheromone trails to converge on near-optimal solutions for CO problems, as detailed in surveys by Blum et al. [27] and Dorigo et al. [28]. Foundational overviews by Dorigo et al. [29] and Fidanova et al. [30] elucidate its mechanics. Recent developments include optimized parameter tuning for TSP [31], bi-heuristic approaches [32], and applications in feature selection [33], multi-objective optimization [34, 35], and path planning [36].

## 3      Preliminary on Ant Colony Optimization for AGV Path Planning

### 3.1    Combinatorial Optimization Problem Formulation

AGV path planning problem can be formulated as a combinatorial optimization task, where the goal is to determine the most efficient path for the AGV from its starting point to its destination within a 3D warehouse. This requires considering various factors, such as distance, cargo attributes, and dynamic obstacles.

Warehouse is represented as a graph $G = (V, E)$, where $V$ denotes the set of vertices, each representing a specific location or waypoint within the warehouse. $V = \{v_0,\ldots, v_n\}$, $v_i = \{x_0, \ldots, x_D\}$, D represents the dimension. $E$ represents the set of paths connecting these vertices. Each edge $e \in E$ has an associated cost $C(e)$. The objective is to find a path $Path = \{v_0,\ldots, v_i\}$ from the starting position of the AGV to the destination vertex $v_i$ that minimizes the total cost:

$$Cost_{total}(P) = \sum_{i,j=v_0,v_1}^{i,j \in P} C(e_{i,j}), \quad (1)$$

where $Cost_{total}(P)$ is the total cost of path, $C(e_{i,j})$ is the cost associated with the edge between vertices $v_i$ and $v_j$ and $P = \{v_0,\ldots, v_i\}$ is the chosen path from the starting position to the destination.

### 3.2 Ant Colony Optimization

A Combinatorial Optimization Problem model defines a pheromone model in the context of ACO. Ant Colony Optimization is a nature-inspired algorithm that emulates the foraging behavior of ants to solve combinatorial optimization problems. In the context of AGV path planning within 3D warehouse environments, ACO offers an efficient approach to navigate complex spaces efficiently.

**Pheromone Model in Ant Colony Optimization.**

ACO simulates the behavior of a colony of ants searching for the shortest path from a source to a destination. Each ant starts at the source location and iteratively constructs a solution by moving through the graph based on two factors:

- Pheromone levels: Ants are attracted to edges with higher pheromone concentrations, which signify favorable paths.
- Heuristic information: In path planning, this can be the inverse of the distance or some other property defined by expert, which encourages ants to choose shorter paths.

In the context of ACO, generally speaking, a pheromone model can be regarded as a construction graph where decision variables serve as nodes and solution components act as edges. Each solution component $s_{i,j}$, which represents the assignment of the value $v_i$ to the decision variable $v_j$, is linked to its pheromone trail $\tau_{i,j}$ and heuristic measure $\eta_{i,j}$. Both $\tau_{i,j}$ and $\eta_{i,j}$ suggest the potential of including $v_{i,j}$ in a solution. Usually, ACO initializes the pheromone trails uniformly and updates them iteratively, while the heuristic measures are predefined and kept constant.

Each ant constructs a path through the graph by probabilistically selecting edges based on pheromone levels and heuristic information. The probability $p_{i,j}^t$ that an ant at vertex $v_i$ selects edge $e_{i,j}$ leading to vertex $v_j$ at time $t$ is given by:

$$p_{i,j}^t = \frac{(\tau_{i,j}^t)^\alpha \cdot (\eta_{i,j}^t)^\beta}{\sum_{k \in N_i}(\tau_{i,k}^t)^\alpha \cdot (\eta_{i,k}^t)^\beta}, \tag{2}$$

here $\tau_{i,j}^t$ is the pheromone level on edge $e_{i,j}$ at time $t$. $\eta_{i,j}$ is the heuristic information predefined by expert for edge $e_{i,j}$, which is usually the inverse of the cost $c_{i,j}$. $\alpha$ and $\beta$ are parameters controlling the relative influence of pheromone and heuristic information, respectively. $N_i$ is the set of neighboring vertices to $v_i$.

**Pheromone Update Mechanism.**

After all ants have completed their paths, the pheromone levels are updated to reinforce successful routes and diminish less favorable ones. The pheromone update rule is:

$$\tau_{i,j}^{t+1} = (1 - \rho)\tau_{i,j}^t + \Delta\tau_{i,j}^t, \tag{3}$$

where ρ is pheromone evaporation rate, $0 < ρ < 1$. $\Delta\tau_{i,j}^t$ is pheromone deposited on $e_{i,j}$ at time $t$, typically calculated as:

$$\Delta\tau_{i,j}^t = \sum_{k=1}^{m} \Delta\tau_{i,j}^k, \tag{4}$$

here $m$ is the number of ants, and $\Delta\tau_{i,j}^k$ is the pheromone contribution from ant, based on the quality of the solution it found. If ant $k$ traversed edge $e_{i,j}$ in its solution, the pheromone update is proportional to the inverse of the path cost $Cost(k)$ of ant $k$:

$$\Delta\tau_{i,j}^k = \frac{Q}{Cost(k)}, \tag{5}$$

where Q is a constant, which is set to 1 in this paper.

In the context of three-dimensional warehouse environments, the AGV pathfinding problem constitutes a combinatorial optimization challenge that can be effectively represented as a graph, with the objective of minimizing path costs. ACO draws inspiration from the foraging behavior of real ants, wherein artificial agents probabilistically construct solutions based on pheromone trails and heuristic information. Upon completion of their respective paths, pheromone levels are updated to reinforce favorable routes, thereby guiding subsequent iterations toward optimal solutions.

## 4     Methodology

AGV path planning problem in 3D warehouses involves determining optimal paths for automated vehicles navigating complex environments. These environments include multi-dimensional cargo properties, dynamic congestion, and a range of operational constraints. To address these challenges, we propose a hybrid methodology combining Ant Colony Optimization and Deep Learning techniques, specifically GNN, and MLP. Our approach is designed to provide more adaptive, accurate, and efficient path planning solutions by leveraging both the strengths of optimization algorithms and deep learning models.

### 4.1    Cargo Modeling and Path Representation

We introduce a multi-dimensional tensor-based cargo modeling system that captures the diverse characteristics of cargo in 3D warehouses. These dimensions include the coordinates, height, size, weight, special handling requirements, and congestion levels. Cargo properties are represented as tensors, with each entry encapsulating a specific aspect of the warehouse's operational state. This modeling approach enables us to better understand the spatial and functional relationships among different cargo items, which improves the accuracy of AGV path planning.

Let $Cargo_i$ represent the multi-dimensional tensor for cargo i where:

$$Cargo_i = [x_i, y_i, z_i, size_i, weight_i, special_i], \tag{6}$$

here $x_i, y_i, z_i$ are the spatial coordinates in a three-dimensional space, $size_i$ is the physical size. $weight_i$ is the mass, and $special_i$ accounts for special characteristics like fragility, perishability or hazardous nature.

### 4.2    Heuristic Function for ACO Integration

Heuristic function in the ACO framework for cargo in 3D warehouses is designed to guide the optimization process based on the cargo and environmental factors. The heuristic matrix $H$ provides guidance by estimating the suitability of each potential move for the AGV. The heuristic is derived from both the cargo characteristics and congestion factors. The heuristic matrix $H$ in this paper is calculated as follows:

$$H_{i,j} = \frac{\gamma \cdot sc_{i,j}}{d_{i,j} + \alpha \cdot size_{i,j} + \beta \cdot wt_{i,j}}, \tag{7}$$

here $size_{i,j}$, $wt_{i,j}$, $sc_{i,j}$ denote the combined the size, weight, special characteristics of $v_i$ and $v_j$. α, β and γ are weighting factors that determine the influence of each attribute on the total cost. $d_{i,j}$ is the Manhattan distance between the locations of cargo items $v_i$ and $v_j$:

$$d_{i,j} = \sum_{k=1}^{D} |v_{i,k} - v_{j,k}|, \tag{8}$$

here D is the number of dimensions of the space and is taken as 3 in this paper.

### 4.3    Integration of Ant Colony Optimization

As described in Section 3.2, ACO is used to explore and exploit the search space of possible AGV paths. The main goal of the ACO framework is to minimize the path cost. In this paper, we propose an innovative algorithm for the path cost of goods in 3D warehouses, introducing the attributes of the goods themselves and the congestion cost. We define the total cost $Cost_{total}$ (P) of the 3D warehouse pickup path P as follows:

$$Cost_{total}(P) = \sum_{i,j \in P} \frac{1}{H_{i,j}} + Con_{i,j}, \tag{9}$$

$$Con_{i,j} = \sum_{i,j \in P} t_{i,j} \cdot \delta(\frac{tc_{i,j}}{cap_{i,j}}), \tag{10}$$

here $H_{i,j}$ is the same as Eq.(7). $t_{i,j}$ is the free flow travel time from cargo $v_i$ to $v_j$. $tc_{i,j}$ is the current traffic flow. $cap_{i,j}$ is the capacity of the edge, which is set to 20 in this article based on Loctek overseas warehouse. δ is adjustment parameter, which is taken as 0.5 in this paper.

After several generations of iterations, the ACO algorithm moves ants in the warehouse map based on pheromone values and heuristic information. The pheromone update equation is shown in Eq.(3), where the cost calculation is illustrated in Eq.(9).

### 4.4 Adaptive Heuristic Modeling for Enhanced Path Optimization in 3D Warehouse Logistics

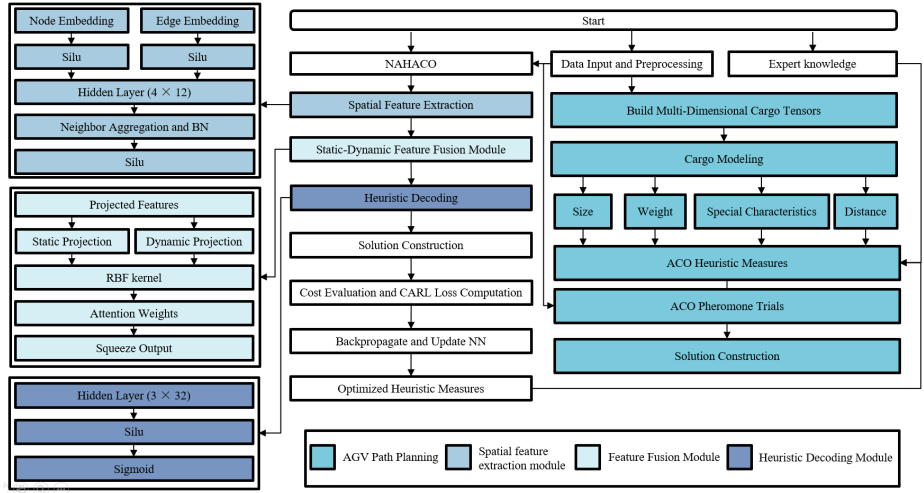

**Fig.1.** The schematic diagrams of NAHACO. It begins with data input and preprocessing, followed by spatial feature extraction via graph-based node and edge embeddings. A static-dynamic feature fusion module then projects refined embeddings into separate subspaces, applies an attention mechanism, and integrates both static and dynamic information. Cargo modeling introduces size, weight, and special handling attributes into the ACO framework, and a decoding MLP generates path heuristics. Finally, the CARL loss function guides backpropagation to produce optimized AGV paths, ensuring efficient and accurate warehouse logistics.

To enhance the ACO-based path planning process, we integrate neural networks to dynamically generate heuristic information. Traditionally, path planning systems often rely on expert-defined heuristics, which are static and limited in their ability to adapt to real-time environmental changes. In contrast, our approach leverages deep learning models to learn the heuristic information from the warehouse's operational data, enabling the system to adapt and generate more effective heuristics. The structure we propose is shown in Fig.1, which is called NAHACO in this paper.

**Spatial Feature Extraction Module**

At the core of our model is a deep learning module that leverages several GNN components to extract and refine spatial relationships from the warehouse layout. The module comprises distinct pipelines for node and edge feature processing, followed by iterative refinement layers that integrate neighborhood information including cargo attributes, spatial distribution information of nodes in the warehouse, and edge features through a mean aggregation mechanism.

Initially, raw node features $x_i$ and edge attributes $a_i$ are projected into higher-dimensional spaces using linear transformations followed by a non-linear activation Silu.

Silu mitigates the issue of dead neurons, ensuring that the network captures the intricate spatial relationships inherent in the warehouse layout. And it is defined as:

$$x_i^{(0)} = Silu(W_v^{(0)} x_i), \tag{11}$$

$$\omega_{ij}^{(0)} = Silu(W_e^{(0)} a_{ij}), \tag{12}$$

here, x represents node features, a represents edge attributes, $\omega$ is edge embeddings, $W_e$ is weight matric to edge embedding and $W_v$ is weight matric of the node.

Over multiple layers ($l = 1, ..., 12$), the node features and edge embeddings are iteratively refined. For the node branch, four parallel linear mappings are applied to the previous node representation:

$$\widehat{\overline{x}}_{i1}^{(l)} = W_{v1}^{(l)} \overline{x}_{i1}^{(l-1)}, \widehat{\overline{x}}_{i2}^{(l)} = W_{v2}^{(l)} \overline{x}_{i2}^{(l-1)}, \widehat{\overline{x}}_{i3}^{(l)} = W_{v3}^{(l)} \overline{x}_{i3}^{(l-1)}, \widehat{\overline{x}}_{i4}^{(l)} = W_{v4}^{(l)} \overline{x}_{i4}^{(l-1)}, \tag{13}$$

neighbor aggregation then consolidates information from neighboring nodes. It plays a key role here. A dedicated neighbor aggregation function collects information from the neighboring nodes by weighting each neighbor's contribution with the transformed edge features:

$$Agg_i^{(l)} = \frac{1}{|N(i)|} (\{ \left( \omega_{ij}^{(l-1)} \overline{x}_i^{(l)} \right) : j \in N(i) \}), \tag{14}$$

here, $N(i)$ and $|N(i)|$ represent the set of all neighboring nodes of node I and the cardinality of this set. This aggregated message is then combined with the intermediate node representation and normalized:

$$x_i^{(l)} = x_i^{(l-1)} + Silu(BN(\overline{x}_i^{(l)} + Agg_i^{(l)})), \tag{15}$$

here, BN is batch normalization. In parallel, edge branch updates each edge embedding by incorporating information from the nodes at its endpoints. Update is given by:

$$\omega_{ij}^{(l)} = \omega_{ij}^{(l-1)} + Silu(BN(W_e^{(l)} \omega_{ij}^{(l-1)} + W_{vi}^{(l)} x_i^{(l-1)} + W_{vj}^{(l)} x_j^{(l-1)})). \tag{16}$$

Overall, our module systematically extracts spatial features by iteratively refining node and edge representations through linear transformations, neighbor aggregation, and batch normalization, thereby capturing the nuanced spatial topology of the warehouse environment.

**Static-Dynamic Feature Fusion Module with Attribute-Aware Attention**

To comprehensively capture the dynamic characteristics of warehouse operations, our module incorporates an attention mechanism into its temporal dependency modeling component. First, a series of spatiotemporal features are extracted from the warehouse data. These features encompass both static spatial features, which signify the intrinsic layout and connectivity of the warehouse, and dynamic temporal features, such as congestion data and real-time location.

These extracted features are combined into two unified spatiotemporal feature matrixes $SFM_s$ and $SFM_d$. To enable the attention mechanism to model temporal dependencies effectively, we map these features into a common subspace using linear transformations:

$$Q_s = SFM_s \cdot W^s, Q_d = SFM_d \cdot W^d, \tag{17}$$

here, $W^s$ and $W^s$ are the learned weight matrices that project the spatiotemporal features into corresponding subspaces. $Q_s$ captures static attributes and $Q_d$ captures dynamic attributes. To model temporal dependencies between these two modalities, we measure the compatibility between $Q_s$ and $Q_d$ using RBF (Gaussian Radial Basis Function) kernel. For target ith cargo, the similarity score is computed as:

$$e_{i,j} = \exp\left(-\frac{\left\|Q_s^i - Q_s^j\right\|^2}{2\sigma^2} + T_{i,j}\right), j \in N(i), \tag{18}$$

where $\sigma$ is a scaling parameter and $T_{i,j}$ is a learnable bias term that captures multi-scale temporal context. $N(i)$ is the set of nodes considered for aggregation for node $i$. The raw scores $e_{i,j}$ are normalized over the neighborhood $N(i)$ to yield the attention weights:

$$A_{i,j} = \frac{e_{i,j}}{\sum_{j \in N(i)} e_{i,j}}, \tag{19}$$

the output representation for each node is computed solely from the static features. In particular, the final output is given by:

$$O_i = Q_d^i + \sum_{j \in N(i)} A_{i,j} \cdot Q_s^j, \tag{20}$$

in effect, it integrates the temporal dynamics via $Q_d$ with complementary spatial context via $Q_s$, enabling the model to capture multi-scale spatiotemporal dependencies more effectively. A subsequent squeeze operation is then applied to restore the representation back to the original input dimensions. Thus, the final enriched representations encapsulate a comprehensive blend of the warehouse's static spatial structure and dynamic operational variations including congestion data while maintaining the required dimensional consistency through the squeeze operation.

**Heuristic Decoding for Path Optimization.**

Processed embeddings from the temporal modeling module are fed into a decoding MLP implemented as a fully connected network that transforms these intermediate representations into heuristic values for path optimization. This decoding MLP is composed of several layers where each hidden layer performs a linear transformation followed by the Silu activation function, and the final layer applies a Sigmoid activation to ensure the outputs are properly scaled for integration into the ACO algorithm.

Let $h^{(i)}$ denotes the input embedding. It is computed as follows:

$$h^{(i)} = \text{Silu}(W^{(i)}h^{(i-1)} + b^{(i)}) \quad i=1, \ldots, L-1, \tag{21}$$

$$\hat{y}^{(i)} = \text{Sigmoid}(W^{(L)}h^{(L-1)} + b^{(L)}), \tag{22}$$

here $W^{(i)}$ and $b^{(i)}$ are the weights and biases of the ith layer. L is the num of layers. Hidden layers employ the Silu activation function because it has self-gating properties that allow it to adaptively scale inputs, which is especially beneficial for capturing subtle patterns in the processed embeddings from the temporal module. The final layer uses the Sigmoid activation function to restrict its output values to the range [0, 1]. The output size is designed to match the number of heuristic predictions for each edge in the graph, enabling dynamically generated heuristics that adapt to fluctuating warehouse conditions and enhance AGV navigation efficiency in complex environments.

### 4.5   CARL Loss Function

Training objective adjusts network parameters to reduce path cost and congestion, thereby boosting AGV performance in warehouse operations. CARL loss function serves as training loss, derived from interactions between network-generated heuristics and ACO path cost computation. Model training is conducted in Loctek testing warehouse environment, using random selection of goods. Loss function follows this equation:

$$L = \frac{1}{n_{ants}} \sum_{i=1}^{n_{ants}} |Cost_i - Cost_{avg}| \cdot \frac{\ln(1+p_i)}{\ln(2)}, \tag{23}$$

where $Cost_i$ is the path cost for the ant i, which is calculated in the same way as Eq.(9). $Cost_{avg}$ is the mean cost over all ants. $p_i$ is the probability of selecting the path taken by the ant i, computed by the ACO algorithm. $n_{ants}$ is the number of ants in the system. logarithmic scaling factor $\frac{\ln(1+p_i)}{\ln(2)}$ modulates this deviation based on the probability of selecting the corresponding path. This logarithmic term is used to stabilize the influence of ants with very low selection probabilities, ensuring that the contribution of each ant to the overall loss remains bounded. The goal of training is to adjust the network parameters to reduce the path costs, as this will improve the overall performance of the AGV in the warehouse.

## 5   Experimentation

### 5.1   Experimental Setup

In this section, we present the experimental setup used to evaluate the proposed multi-dimensional AGV path planning approach that combines ACO with advanced neural networks in a 3D warehouse environment. Our approach leverages a tensor-based modeling system to handle cargo characteristics, a deep learning model designed in Section 4 for parameter prediction, and a congestion-aware loss function in ACO for dynamic path cost adjustment. Our goal is to demonstrate the effectiveness of our approach in optimizing AGV path planning in a real-world large-scale warehouse instance while considering key factors such as congestion, cargo size, weight, and special handling requirements. The number of training epochs is 100000 with the CARL loss function.

Our experiments were conducted on a workstation equipped with one Intel i7-10875H CPU, one Nvidia GeForce 2080 SUPER GPU and Ubuntu 20.04. This hardware setup ensured efficient computation during both the deep learning model training and the iterative optimization processes of the ACO framework.

We compare the proposed model with traditional and state-of-the-art path planning methods in two phases: (1) simulating a large-scale TSP (Traveling Salesman Problem) for baseline comparison; (2) applying it to a real warehouse instance with various operating conditions under the Loctek test warehouse. The evaluation is based on the following four key performance indicators:

- **Time:** This indicator records the computational time in seconds required to reach a solution. It is crucial for assessing the efficiency and scalability of the approach, particularly in large-scale or real-time applications.
- **Cost (Objective Function Value):** This indicator represents the computed total cost of the solution. In our TSP simulation, it is defined in the same way as in Eq.(12). A lower total cost indicates a more optimal solution.
- **Gap**: This metric measures the percentage deviation of the obtained total cost from the LKH-3 solution, which is used as the baseline.
- **Con:** A lower Congestion value indicates a more effective routing strategy that minimizes traffic bottlenecks in the warehouse environment, which is defined in the same way as in Eq.(10).

### 5.2   Large-Scale TSP Simulation

In the first stage, we evaluate NAHACO on large-scale TSP benchmarks using datasets with 200, 500, and 1000 nodes generated with PyTorch. In this controlled environment, each node represents a key waypoint with attributes defined in our model. The objective is to reduce the overall path cost by considering both travel distance and congestion penalties with $\delta$ set to 0.

**Tab.1.** Comparison results on TSP200, TSP500, and TSP1000 (with LKH-3 as the baseline solution).

| Method | TSP200 | | | TSP500 | | | TSP1000 | | |
|---|---|---|---|---|---|---|---|---|---|
| | Time | Cost | Gap | Time | Cost | Gap | Time | Cost | Gap |
| LKH-3 [37] | **1.3s** | 7.3 | - | **2.8s** | 17.3 | - | 12.1s | 24.3 | - |
| ACO | 3.2s | 13.4 | 6.4% | 8.6s | 20.4 | 7.6% | 56.3s | 24.9 | 9.2% |
| SO [11] | 7.4s | 8.8 | 1.9% | 15.3s | 16.8 | 3.1% | 27.4s | 23.8 | **3.7%** |
| RRT-ACO [38] | 1.3s | 8.4 | 1.7% | 4.6s | 15.4 | 2.8% | 13.2s | 24.3 | 4.2% |
| DeepACO [39] | 1.6s | 7.7 | **1.2%** | 4.3s | 17.7 | 2.8% | **11.5s** | 23.9 | 3.8% |
| DIMES [8] | 4.5s | 7.8 | 1.3% | 10.3s | 17.8 | 2.7% | 31.7s | 24.4 | 4.7% |
| NAHACO | **1.3s** | **5.5** | **1.2%** | 5.4s | **13.3** | **2.5%** | 12.8s | **21.5** | **3.7%** |

Tab.1 compares NAHACO with several methods. For TSP200, LKH-3 achieves a total cost of 7.3 in 1.3 seconds, while NAHACO achieves a cost of 5.5 in 1.3 seconds, outperforming both ACO and DIMES [8]. In the TSP500 scenario, NAHACO records a total cost of 13.3 in 5.4 seconds, which is lower than those obtained by ACO and DIMES and has a shorter runtime than SO [11]. For TSP1000, NAHACO achieves a cost of 21.5 in 12.8 seconds. This is a 11.5% reduction in cost compared to LKH-3's [37] cost of 24.3 and shows a 60% improvement in the optimality gap over ACO.

Results show that NAHACO maintains high solution quality while offering competitive computational efficiency. Its runtime grows linearly with problem size, unlike ACO's exponential increase. Overall, NAHACO demonstrates efficient performance and scalability for large-scale TSP optimization, effectively integrating deep learning to enhance heuristic search strategies.

### 5.3  Real-World Warehouse Instance

**Tab.2.** Comparison results on the Loctek test warehouse with 100, 200 and 500 cargos randomly selected on the shelves.

| Method | 100 Cargos | | | 200 Cargos | | | 500 Cargos | | |
|---|---|---|---|---|---|---|---|---|---|
| | Time | Cost | Con | Time | Cost | Con | Time | Cost | Con |
| LKH-3 [37] | **2.1s** | 19.3 | 29.4 | **5.3s** | 27.4 | 39.4 | **14.7s** | 37.7 | 47.3 |
| ACO | 7.6s | 35.9 | 59.7 | 32.6s | 48.1 | 62.7 | 73.8s | 65.2 | 89.1 |
| SO [11] | 10.4s | 16.8 | 23.1 | 23.1s | 28.3 | 34.6 | 52.6s | 41.5 | 64.2 |
| RRT-ACO [38] | 3.4s | 14.7 | 27.2 | 9.3s | 25.2 | 39.2 | 23.8s | 34.1 | 48.7 |
| DeepACO [39] | 3.7s | 14.1 | 31.3 | 7.6s | 25.8 | 41.6 | 19.5s | 34.9 | 49.6 |
| DIMES [8] | 8.9s | 15.6 | 30.4 | 21.4s | 34.2 | 40.1 | 51.7s | 39.6 | 50.6 |
| NAHACO | 4.3s | **10.3** | **15.3** | 13.4s | **18.9** | **23.4** | 35.4s | **24.3** | **28.2** |

In the second stage, we test our method on a real warehouse instance from the Loctek test facility. In this experiment, we create a comprehensive 3D warehouse environment with 100, 200, and 500 cargos selected at random from the shelves. Each cargo is represented by a multi-dimensional tensor as defined in Eq.(6). The cost function is specified in Eq.(12) and the congestion function in Eq.(10), with δ set to 0.2.

The results are shown in Tab.2. For 100 cargos, NAHACO reduces cost by 39% compared to SO and lowers congestion by 34%. For 200 cargos, NAHACO achieves a 25% lower cost than RRT-ACO [38] and cuts congestion by 63% compared to ACO. In the 500-cargo scenario, NAHACO records the lowest cost and congestion values among all methods. It reduces cost by 39% relative to DIMES and improves congestion by 68% compared to ACO, even though its computation time is slightly longer. In contrast, ACO suffers from exponential increases in congestion as cargo count grows, and SO's cost rises by 147% when scaling from 100 to 500 cargos.

While NAHACO maintains high solution quality and scalability for large-scale TSP optimization, its runtime is slightly slower than that of RRT-ACO and DeepACO [39] due to several intertwined factors. Its integration of deep learning for dynamic heuristic adjustment, although beneficial for enhancing search strategies, introduces additional computational overhead not present in the lighter reinforcement learning frameworks of RRT-ACO and DeepACO. Moreover, the hybrid architecture of NAHACO, which combines traditional ant colony optimization with deep neural network components, necessitates extra processing steps and a more complex coordination of algorithms. This intricate balance between exploration and exploitation, along with the continuous adaptation of heuristic parameters based on learned insights, results in a marginal increase in runtime. Despite this, the trade-off ensures efficient performance and consistently high-quality solutions for complex, large-scale TSP instances.

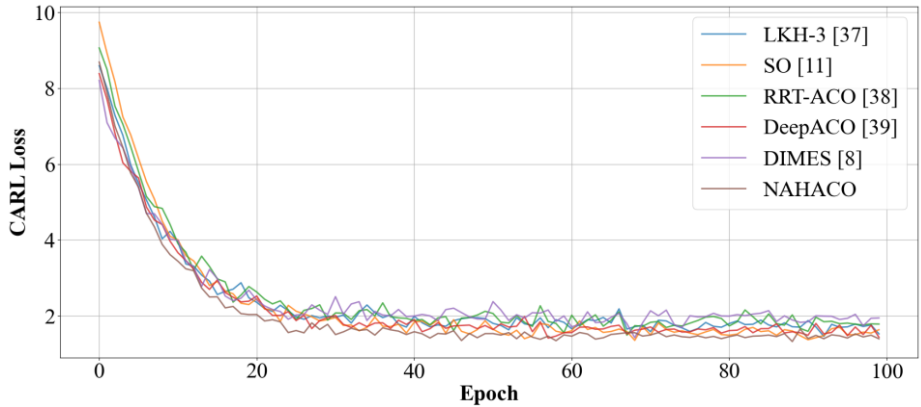

**Fig.2**.Training CARL loss curves over 100 epochs.

Training CARL loss curves is shown in Fig.2. NAHACO achieves the lowest final CARL loss among all compared methods, indicating its superior solution quality. Furthermore, it converges faster and more stably than other approaches, ensuring efficient performance for large-scale warehouse picking route planning. These results demonstrate that NAHACO adapts well to dynamic warehouse environments and efficiently optimizes resource allocation, making it an efficient solution for real-world warehouse automation.

## 6   Conclusion

In this work, we introduce NAHACO, a multi-dimensional AGV path planning framework merging Ant Colony Optimization with neural networks. It employs a tensor system for cargo attributes and a hybrid architecture with Spatial Feature Extraction, Temporal Dependency Modeling, and Heuristic Decoding modules to enable accurate ACO parameter prediction and real-time path cost adjustments. In the TSP benchmarks, NAHACO consistently achieves lower total costs compared to LKH-3 and other methods—with cost reductions ranging from approximately 11.5% to nearly 25%. In Loctek's warehouse tests, NAHACO cuts cost by up to 47% and reduces congestion by up to 48% compared to LKH-3. NAHACO also maintains competitive computational efficiency. These results prove that NAHACO is effective solution for complex logistics.

**Acknowledgments.** This work was partly supported by Ningbo Youth Science and Technology Innovation Leading Talent Project (2024QL044). The authors gratefully acknowledge the support and resources provided by Zhejiang University, whose academic environment and research infrastructure have been instrumental in the development and execution of this work. We also extend our sincere thanks to Loctek for granting access to their test warehouse facilities, providing critical datasets, and offering valuable industrial insights that have significantly enriched our experimental evaluations. Their contributions have been vital to the success of this research.